\documentclass[journal]{IEEEtran}
\usepackage[utf8]{inputenc}
\pdfoutput=1
\usepackage[T1]{fontenc}
\usepackage[english]{babel}
\usepackage{color}
\usepackage{multirow}
\usepackage{multicol}
\usepackage{hyperref}
\hypersetup{
    colorlinks=true,
    linkcolor=blue,
    filecolor=magenta,      
    urlcolor=black,
}
\usepackage{amsmath,amsxtra,amssymb,latexsym, amscd,amsthm}
\usepackage[pdftex]{graphicx}
\usepackage[mathscr]{eucal}
\usepackage{mathtools}
\usepackage{balance}

\makeatletter

\providecommand{\tabularnewline}{\\}

\ifCLASSINFOpdf
\else
\fi
\usepackage{caption}
\makeatother

\usepackage{babel}
\begin{document}
\title{Anomaly detection using prediction error with Spatio-Temporal Convolutional LSTM}

\author{\IEEEauthorblockN{Hanh T. M. Tran\IEEEauthorrefmark{1}, David Hogg \IEEEauthorrefmark{2}}
	
\IEEEauthorblockA{\IEEEauthorrefmark{1} The University of Danang, University of Science and Technology, Vietnam} \\
\IEEEauthorblockA{\IEEEauthorrefmark{2} School of Computing, University of Leeds, United Kingdom}}

\maketitle
\begin{abstract}
In this paper, we propose a novel method for video anomaly detection motivated by an existing architecture for sequence-to-sequence prediction and reconstruction using a spatio-temporal convolutional Long Short-Term Memory (convLSTM). As in previous work on anomaly detection, anomalies arise as spatially localised failures in reconstruction or prediction. In experiments with five benchmark datasets, we show that using prediction gives superior performance to using reconstruction. We also compare performance with different length 
input/output sequences. Overall, our results using prediction are comparable with the state of the art on the benchmark datasets. 

\end{abstract}



\section{Introduction}
Automatically detecting abnormal events in video has been widely studied in recent years due to its broad range of applications, including wide-area surveillance and health monitoring. This problem is different from event detection where the event is clearly defined, since an anomaly is by definition unknown in advance and may arise from unfamiliar activities or activities in unfamiliar contexts.

The standard approach to anomaly detection has been to learn spatio-temporal models of normal activity using hand-crafted features \cite{kim2009observe,mahadevan2010anomaly,cong2011sparse,lu2013abnormal,adam2008robust,wanganomaly} or deep feature representations \cite{tran2017anomaly,xu2015learning}. An abnormality is detected when spatio-temporal patterns are observed that do not conform to the model of normality.
Many different low-level features using dense optical flow (e.g., histograms \cite{adam2008robust}, MHOF \cite{cong2011sparse} ) and other patterns of spatio-temporal gradient \cite{lu2013abnormal} have been used in the past. A model of normality is learned using these features extracted from training data and then used to determine numerical abnormality scores in test data. The model may be of several different kinds, including probabilistic models (e.g, mixture of probabilistic PCA \cite{kim2009observe}, mixture of dynamic texture \cite{mahadevan2010anomaly}), domain based (e.g, one-class SVM \cite{wanganomaly}), sparse coding \cite{cong2011sparse} and Sparse Combination Learning (SCL) \cite{lu2013abnormal}. All of these methods have been used for anomaly detection and localization within the image frame. 

Recently, deep learning architectures have been successfully applied in many computer vision tasks including video anomaly detection. A key advantage of deep learning methods is that they can learn feature representations directly from training data without prior definition. For example, this can be done in an unsupervised manner using auto-encoders \cite{xu2015learning,tran2017anomaly,tran2018anomaly}. A stacked de-noising autoencoder can be used to learn appearance and motion features for anomaly detection \cite{xu2015learning}. A Winner-take-all sparsity constraint combined within the autoencoder has been shown to produce flow-features that are more discriminative for a one class SVM \cite{tran2017anomaly} that is trained separately on the compressed representations learnt by the autoencoder.

\begin{figure*}[ht]
\begin{center}
\includegraphics[width=0.99\textwidth]{{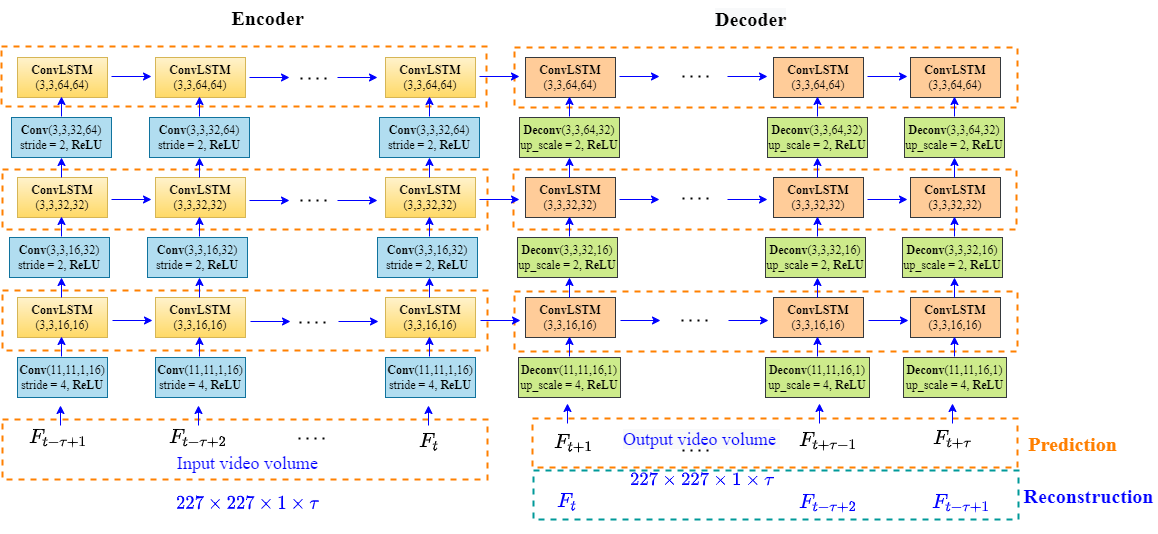}}
\end{center}
\vspace{-20pt}
  \caption{The encoding-decoding structure used for future prediction or reconstruction with video volumes of $\tau$ frames.}
\label{fig:model}
\end{figure*}
\begin{figure}[!h]
\begin{center}
\includegraphics[width=0.4\textwidth]{{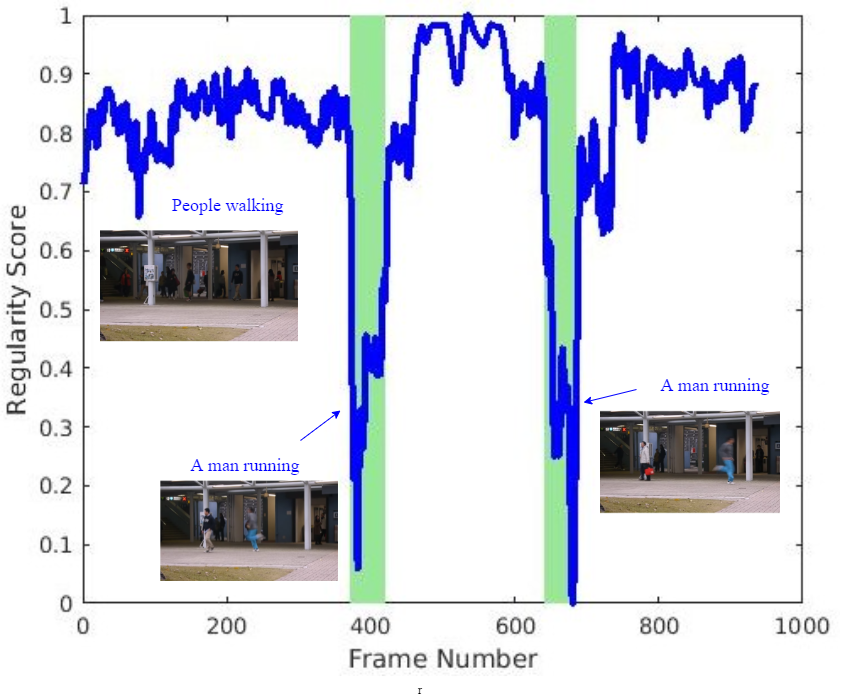}}
\end{center}
\vspace{-20pt}
   \caption{The regularity score of video sequence of the CUHK Avenue dataset~\cite{lu2013abnormal}. The score decreases when an anomaly (a running man) appears on the scene.}
\label{fig:result_demo}
\end{figure}

End-to-end deep learning approaches have also been proposed for anomaly detection \cite{hasan2016learning,ravanbakhsh2017training,zhao2011online,liu2018future,Nguyen_2019_ICCV}. 
The convolutional autoencoder (convolutional AE) can be used to learn a model of normality from video, then reconstruction error \cite{hasan2016learning,ravanbakhsh2017training,zhao2011online} or prediction error \cite{liu2018future,Nguyen_2019_ICCV} provide a local measure for anomaly detection. 
A Generative Adversarial Network (GAN) can be employed to generate a normal distribution over some datasets \cite{ravanbakhsh2017training,liu2018future} by jointly optimising with a discriminator that competes to distinguish what is a real normal sample from what is a generated one. 
The motion dynamic can be learnt using a multi-channel approach, fusing appearance and motion information, and a cross-channel task, forcing the generator to transform raw-pixel data into motion information and vice versa \cite{ravanbakhsh2017training} or using FlowNet combined with U-net for a single frame prediction\cite{liu2018future}. 
The combination of a convolutional autoencoder and U-net has also been used to build two-stream network with a shared encoder in which one decoder is for a single frame reconstruction and one is for translating an image to optical flow \cite{Nguyen_2019_ICCV}. 
Another Spatio-Temporal autoencoder has been proposed for video anomaly detection \cite{zhao2017spatio}. The results show that applying 3D convolution in the encoder and 3D deconvolution in the decoder helps to enhance the capability of extracting motion patterns over the temporal dimension.

A memory module is proposed into the AE to address these problems \cite{gong2019memorizing,park2020learning}. The encoder inputs a normal video frame and extracts feature maps. The encoding features are then used to retrieve prototypical normal patterns in the memory items and to update the memory. Then the feature maps and aggregated memory items  are fed into the decoder for reconstructing the input video frame or predicting the next frame. Using cosine similarity and the softmax function for matching probability between incoming encoding features and memory items, the global memory can be read and written to. Since normal patterns in training and testing sets may be different, the memory items are updated during training and testing time, with the use of a predefined threshold to prevent updating on anomaly patterns \cite{park2020learning}. However, it is impossible to find an optimal threshold to distinguish between normal and abnormal patterns under various scenarios. Meta-learning methodology is introduced into a Dynamic Prototype Unit (DPU) to learn prototypes for encoding normal dynamics and to enable the fast adaption capacity to a new scene with only a few training frames \cite{lv2021learning}. As in previous work \cite{park2020learning}, the DPU inputs the encoding feature maps, which are outputs of the encoder part of U-net, to generate a pool of dynamic prototypes. However it is trained in a fully differential attention manner in which attention mapping functions are implemented as fully connected layers and updated using gradient descent style. After training the AE backbone using only frame prediction loss, the DPU module is trained with the meta-training phase using frame pairs sampled from videos of diverse scenes. In the testing phase, in order to adapt the model to a new scene, the first few frames of the sequence in this scene are used to construct K-shot input-output frame pairs. The results show that the DPU is more memory-efficient than the memory module in previous work \cite{gong2019memorizing,park2020learning}.

Another approach to learning regular spatio-temporal patterns is to use a convolutional LSTM \cite{chong2017abnormal,luo2017remembering}. The motivation is that reconstruction over a longer duration using the memory of the LSTM should capture more complex flow patterns. 
The convolutional network is used to encode each frame, then feeding these encoding tensors to Convolutional LSTMs to memorize the change of the appearance which corresponds to motion information \cite{luo2017remembering}. Two Deconvolutional Networks (DeconvNet) are used, one for reconstructing past frames and to identify whether an anomaly occurs; and one for reconstructing the current frame. Thus the reconstruction error is an indicator of the change in appearance or motion. The temporal unit in \cite{chong2017abnormal,luo2017remembering} is applied on the final spatial stage, which encodes high level representations. Interleaving RNNs between spatial convolution layers has recently been shown to improve performance on precipitation now-casting \cite{shi2017deep}. The model can learn temporal information on hierarchical spatial representations from low-level to high-level. 
In our work, we adopt the same architecture, except that we remain with convolutional LSTMs instead of the complex trajGRU RNN \cite{shi2017deep}.
Our results show a comparable level of performance to the state of the art on benchmark datasets with fewer model parameters than state of the art models. Moreover, using prediction gives better performance than reconstruction. Finally, performance varies as expected with different prediction windows.  

\section{Architecture}
\vspace{-5pt}
Figure \ref{fig:model} illustrates the encoding-decoding structure for future prediction or reconstruction, motivated by earlier work \cite{shi2017deep} and adapted for anomaly detection.
At each time step, the network takes a video volume of $\tau$ video frames $F_{t-\tau+1}, ..., F_t$, and generates an output volume of the same size, predicting the future $F_{t+1},...,F_{t+\tau}$ or reconstructing the input in reverse order $F_{t},...,F_{t-\tau+1}$.
\subsection{Encoding-decoding model}
The structure consists of two networks, an encoding network and a decoding network (Fig. \ref{fig:model}). The encoder contains three convolutional layers, each followed by leaky ReLU with negative slope equal to 0.2 \cite{maas2013rectifier}. In order to do down-sampling, we use all three convolutional layers with stride. The strided convolution allows the network to learn its own spatial down-sampling. Similarly, three deconvolution layers are used in the decoder to learn its own spatial up-sampling. 
The goal of temporal encoding is to capture and compress changes due to motion in the input sequence into encoding hidden states that allow the decoder to reconstruct the input or predict the future. 
Spatio-temporal LSTM cells \cite{xingjian2015convolutional} are employed as a temporal encoder/decoder. At each time $t$, the convolutional LSTM (convLSTM) module receives as input a new video frame after projection in the spatial feature space. This is used together with the memory content and output of the previous step $t-1$ to compute new memory activations. 
Interleaving multiple convLSTMs between convolutional layers helps the model learn spatio-temporal dynamic information at different levels. The high level states capture global spatial-temporal representations while the lower level states retain the detail of local spatio-temporal representations. 
After the last frame is read, the decoding LSTMs take corresponding states from the encoder as their initial states and output an estimate for the target sequence (Fig. \ref{fig:model}). The low-level states are combined with the up-sampling outputs as the initial states and inputs of decoding LSTMs helps to aggregate low-level information to the up-scaling data stream. Therefore, the output contains details on both background and object (Fig. \ref{fig:recons-predict}). 
\subsection{Input data layer}
\begin{table*}[!htpb]
\caption{Performance comparison with the state of the art.}
\label{table:comp1}
\begin{center}
\begin{tabular}{l p{1.9cm} p{1.9cm} p{1.5cm} p{1.5cm} p{1.5cm} }
\hline
\multirow{3}{*}{{\bf Method}} & \multicolumn{5}{c}{AUC/EER (\%)} \\
\cline{2-6}
& UCSDPed1 & UCSDPed2 & CUHK Avenue & Subway Entrance & Subway Exit\\
\hline
Conv-WTA\cite{tran2017anomaly}  & $91.6/\underline{14.8}$ & $95/$\(\mathbf{9.5}\) & $81/26.5$ & - & - \tabularnewline
AMDN\cite{xu2015learning}   & $\underline{92.1}/16$ & $90.8/17.1$  & - & -  \tabularnewline
GAN \cite{ravanbakhsh2017training}   & - & $93.5/15.6$ & - & - & - \\
Conv-AE \cite{hasan2016learning}  & $81/27.9$ & $90/21.7$ & $70.2/25.1$ & \(\mathbf{94.3}\)$/26.0$ & $80.7/9.9$  \\
ST-AE\cite{chong2017abnormal}  & $89.9/$\(\mathbf{12.5}\) & $87.4/\underline{12.0}$ & $80.3/$\(\mathbf{20.7}\) &$84.7/\underline{23.7}$ & $\underline{94.0}/\underline{9.5}$\\ 
Past-Current-LSTM \cite{luo2017remembering}& $75.5/-$ & $88.1/-$ & $77/-$ &$\underline{93.3}/-$ & $87.7/-$ \\ 
STAE-3D\cite{zhao2017spatio} & \(\mathbf{92.3}\)$/15.3$ & $91.2/16.7$ & $77.1/33.8$ &$-$ & $-$ \\ 
FlowNet-Unet-GAN \cite{liu2018future} & $83.1/-$ & $95.4/-$ & $85.1/-$ &$-$ & $-$ \\ 
Two-streams AE \cite{Nguyen_2019_ICCV} & - & $94.1/-$ & $83.3/-$ &$-$ & $-$ \\
MemAE \cite{gong2019memorizing} & - & $96.2/-$ & $86.9/-$ &$-$ & $-$ \\
LMN \cite{park2020learning} & - & \(\mathbf{97}\)$/-$ & $\underline{88.5}/-$ &$-$ & $-$ \\
MPD* \cite{lv2021learning} & $83.2/-$ & $95.1/-$ & $84.0/-$ &$-$ & $-$ \\
MPD \cite{lv2021learning} & $85.1/-$ & $\underline{96.9}/-$ & \(\mathbf{89.5}\)$/-$ &$-$ & $-$ \\
\hline
Ours (prediction)  & $80.8/25.1$ &$92.3/14.4$ & $84.8/\underline{22.4}$ & $90.2/$\(\mathbf{15.9}\) & \(\mathbf{95}\)$/$\(\mathbf{8}\) \\
\hline
\end{tabular}
\end{center}
\end{table*}
The input to the model is a video volume consisting of $\tau$ consecutive frames. Each frame is extracted from raw video, converted to a gray-scale image and resized to $227 \times 227$. The pixel values are scaled to the range $[0,1]$. We stack $\tau$ frames in the $4^{th}$ dimension into video volumes and use them as the input of size $227\times227\times1\times\tau$ to the encoder. 
Following \cite{hasan2016learning}, we generate more video sequences by concatenating frames with skipping strides of 1, 2 and 3, thereby simulating faster motion patterns. Although speed can be important in anomaly detection, we still carry out this augmentation to minimise over-fitting and to have a fair comparison with \cite{hasan2016learning,chong2017abnormal}. Unlike \cite{hasan2016learning}, we do not stack precomputed optical flow into our input volume, in the expectation that the network can learn the necessary patterns of motion.
\section{Training}
The weights \(\mathbf{W}_l\) and biases \(b_l\) of each layer \(l\) are learned by minimizing the regularized least squares error: 
\begin{equation} \label{eq:2}
\frac{1}{2N\tau}\sum_{n=1}^N \|\theta_n - \hat{\theta}_n\|_2^2 + \frac{\lambda}{2}\sum_{l} \|\mathbf{W}_l\|_2^2
\end{equation}
where $\hat{\theta}_n$ is the predicted frame sequence (or the reconstructed frame sequence) from the model and $\theta_n$ is the target sequence. The first term is the prediction error (or the reconstruction error) and the second term is to regularize the weights. $\lambda$ is a hyper-parameter used to balance the importance of two terms. 

The weights in each convolutional layer are initialized from a zero-mean Gaussian distribution with standard deviation calculated from the number of input channels and the spatial filter size of the layer~\cite{he2015delving}. This is a robust initialization method that particularly considers the rectifier nonlinearities. We initialize the weights for convLSTM using a zero-mean Gaussian distribution with a fixed standard deviation of 0.01. The biases for all layers are initialized to zero. The input-to-hidden and hidden-to-hidden convolutional filters in the convLSTM cell are the same size.

\subsection{Anomalous event detection} \label{anomaly detection}

The Adam \cite{kingma2014adam} method is used to optimize the error in Eq. \ref{eq:2} with batch size \(N=4\), momentum of $0.9$ and $0.999$, and weight decay  \(\lambda = 5 \times 10^{-4}\) ~\cite{krizhevsky2012imagenet}. We train a network separately on each dataset so that the model learns the specific normal patterns. An event may be normal in one dataset but abnormal in another. For example, people going towards the turnstile to enter the subway station is normal in the Subway Entrance dataset but abnormal in the Subway Exit dataset. We start training the model with a learning rate of $10^{-4}$. After 80 epochs, we stop training and use the model for anomaly detection.
\section{Regularity score for anomaly detection}
Once the model is trained, the prediction error between each output frame $\hat{F}_i$ and the target frame $F_i$ in the video sequence is computed, then errors of all $\tau$ frames are summed up  to form the prediction error for a volume as follows:
\begin{equation} \label{eq:3}
e(t) = \sum_{i=t+1}^{i =t+\tau}||\hat{F}_{i} - F_{i}||_2
\end{equation}
The prediction error then is normalized to compute a regularity score $s(t)$ of a testing volume as follows \cite{hasan2016learning}:
\begin{equation} \label{eq:4}
s(t) = 1 - \dfrac{e(t) - min_{t'}e(t')}{max_{t'}e(t')}
\end{equation}
where $min_{t'}e(t')$ and $max_{t'}e(t')$ are calculated over the prediction errors of all volumes in the same test video. If the regularity score $s(t)$ is less than a threshold, the corresponding test volume is abnormal. 

We also use the same architecture for reconstruction in our experiments. Instead of using the next $\tau$ frames as the target sequence, we use the input sequence in reverse order as the target. Replacing the target sequence in Eq. \ref{eq:3}, we obtain the reconstruction error and use it for anomaly detection with the reconstruction model.
\section{Experiments}
Our method is evaluated both quantitatively and qualitatively. We modify and use Caffe\cite{jia2014caffe} for all our experiments. Code and trained models are available at \textit{\url{https://github.com/t2mhanh/convLSTM_Prediction_AnomalyDetection}}. 
\vspace{-10pt}

\begin{figure}[!htpb]
\begin{center}
\begin{tabular}{c}
\includegraphics[width=0.47\textwidth]{{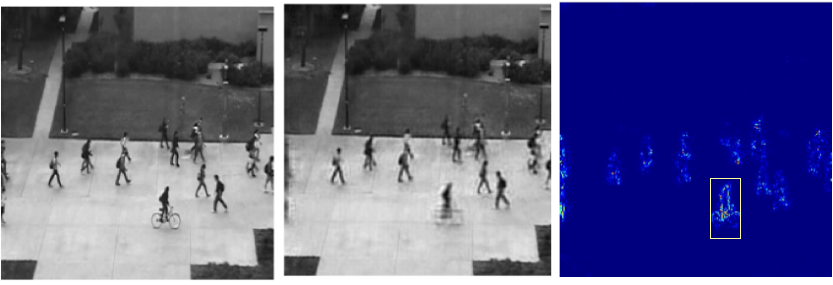}}\\
Reconstruction (error \( e(t) = 21.57\)) - UCSDPed2 - biker\\
\includegraphics[width=0.47\textwidth]{{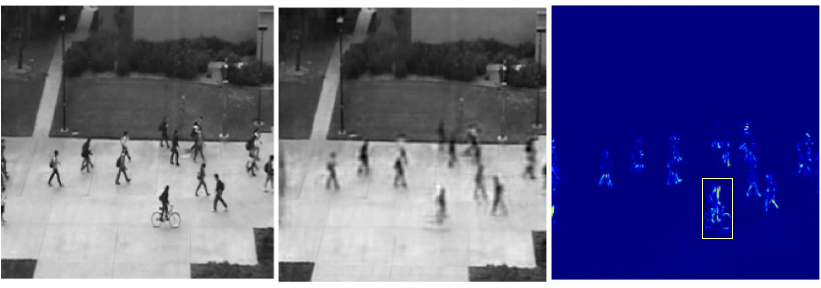}}\\
Prediction (error \( e(t) = 41.03\)) - UCSDPed2 - biker \\
\includegraphics[width=0.47\textwidth]{{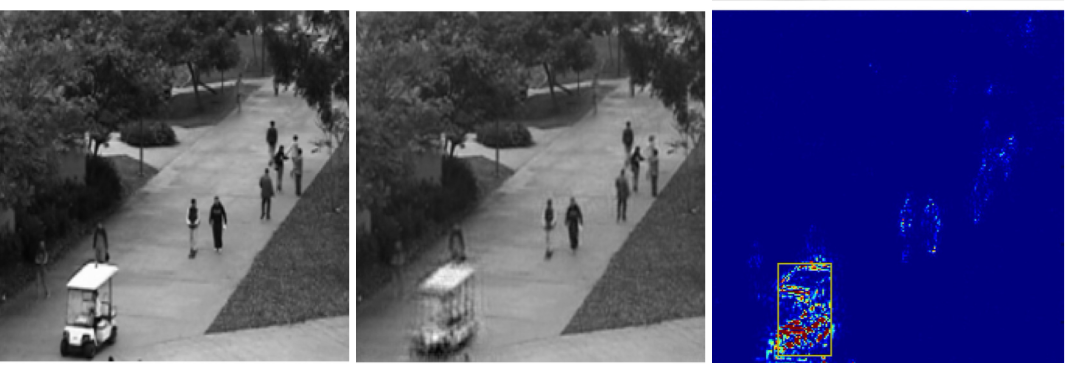}}\\
Reconstruction (error \( e(t) = 30.28\)) - UCSDPed1 - car  \\
\includegraphics[width=0.47\textwidth]{{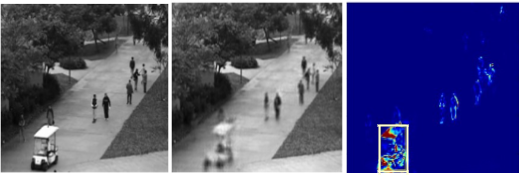}}\\
Prediction (error \( e(t) = 66.25\))- UCSDPed1 - car  \\
\includegraphics[width=0.47\textwidth]{{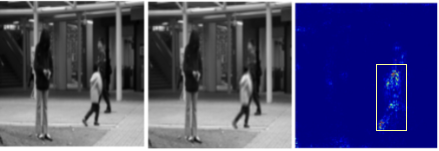}}\\
Reconstruction (error \( e(t) = 11\)) - CUHK Avenue - running  \\
\includegraphics[width=0.47\textwidth]{{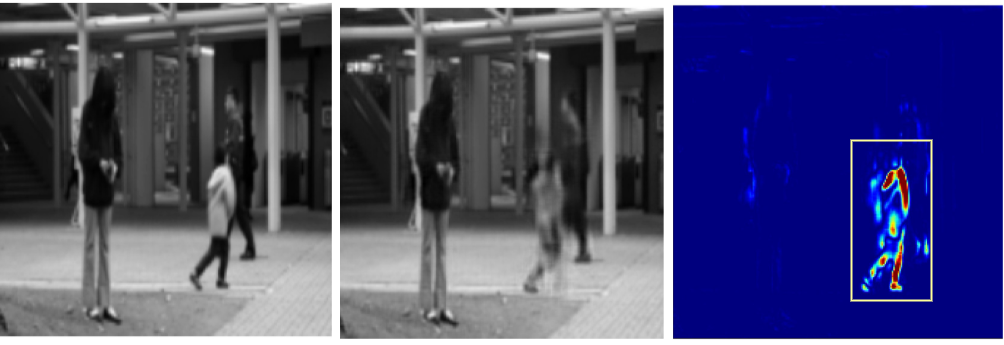}}\\
Prediction (error \( e(t) = 59.81\)) - CUHK Avenue - running  \\
\end{tabular}
\end{center}
\vspace{-10pt}
   \caption{Prediction and reconstruction of third frame out of $5$ (middle), compared to target frame (left); accumulated per-pixel error over $5$ frames as blue-green-red colour map (right). Ground truth anomalies shown as rectangles. Taken from UCSDPed2, UCSDPed1 and CUHK Avenue. Best viewed in color} 
\label{fig:recons-predict}
\end{figure}
\setlength\intextsep{0pt}
\vspace{0pt}


\subsection{Datasets}
Our models are trained on five of the most commonly used datasets for anomaly detection: UCSD (UCSDPed1 and UCSDPed2)\cite{mahadevan2010anomaly}, CUHK Avenue \cite{lu2013abnormal}, Subway (Entrance and Exit) \cite{adam2008robust}. The UCSD and CUHK datasets have separate training videos which contain mostly normal events. The first 12 minutes of Subway Entrance and the first 5 minutes of Subway Exit are used for training. 
\subsection{Anomalous event detection}
Two performance metrics are employed for evaluation and comparison with state of the art results: Equal Error Rate (EER) and Area Under the ROC Curve (AUC). The regularity score of each volume determines whether it is normal or abnormal. We follow the intuition that testing video volumes containing normal events generate high regularity scores (Eq. \ref{eq:4}) since they are similar to training data. A testing video sequence containing an anomaly gives a lower score. Setting different thresholds on the regularity score, volumes are classified into those that contain an anomaly and those that do not. These predictions are compared with ground-truth to give the equal error rate (EER) and area under the curve (AUC) of the resulting ROC curve (TPR versus FPR) generated by varying an acceptance threshold.
Good performance has a low EER and high AUC. 

Table \ref{table:comp1} shows that the model trained for prediction performs comparably to state of the art results. Performance on UCSDPed1 is relatively poor, whilst for CUHK Avenue, the AUC is better than most methods, except FlowNet-Unet-GAN \cite{liu2018future}, MemAE \cite{gong2019memorizing}, LMN \cite{park2020learning}, MPD \cite{lv2021learning}. However, MemAE \cite{gong2019memorizing}, LMN \cite{park2020learning} and MPD \cite{lv2021learning} have more parameters than our models which is shown in table \ref{table:speed}.

\begin{table}[!h]
\caption{Comparison of AUC/EER with different models. $\tau$ is the number of frames in an input sequence and a target sequence.} 
\label{table:difmodels}
\begin{center}
\begin{tabular}{l p{.857cm} p{1.4cm} p{1.4cm} p{1.2cm}}
\hline
\multirow{3}{*}{{\bf Method}} & & \multicolumn{3}{c}{AUC/EER (\%)}\\
\cline{3-5}
& & UCSDPed1 & UCSDPed2 & CUHK Avenue\\
\hline
\multicolumn{2}{l}{Reconstruction}  & $75.6/28.9$ &$87.5/17.1$ & $81.4/26.1$\\
\hline
\multirow{3}{*}{Prediction} &
 $\tau=2$ & $78.3/27.1$ & $86.1/21.1$ & \(\mathbf{85.1}\)$/22.5$   \\
\cline{2-5}
& $\tau=5$ & \(\mathbf{80.8/25.1}\) & \(\mathbf{92.3/14.4}\) & $84.8/$\(\mathbf{22.4}\) \\
\cline{2-5}
& $\tau=8$ & $79/26.5$ & $89.6/18.5$ & $83.2/23.2$   \\
\hline
\end{tabular}
\end{center}
\end{table}

Table \ref{table:difmodels} shows the results when different models are used. In the table, ``Reconstruction'' is for a model trained for reconstructing a sequence of 5 frames and ``Prediction'' is for models trained to predict $\tau$ frames. The model trained for future prediction gives better results than the reconstruction model. This may be because prediction will always try to draw back to normality, whereas reconstruction works from pre-sight of an anomalous sequence. The quality comparison between reconstruction and prediction is shown in Figure \ref{fig:recons-predict}.

\begin{table}[!h]
\caption{Comparison of model complexity and testing speed.} 
\label{table:speed}
\begin{center}
\begin{tabular}{l| p{2.4cm}| p{0.7cm} }
\hline
Methods & Parameters (M) & FPS \\
\hline
Conv-AE \cite{hasan2016learning} & $8.4$ & $-$\\
ST-AE \cite{chong2017abnormal} & $1.1$ & $-$ \\
STAE-3D \cite{zhao2017spatio} & $0.5$ & $-$\\
MemAE \cite{gong2019memorizing}  & $6.2$ & $45$\\
LMN \cite{park2020learning} & $15.0$ & $67$\\ 
\hline
Ours & $0.85$ & $75$  \\
\hline
\end{tabular}
\end{center}
\end{table}

The number of model parameters for the method against
different end-to-end trainable models in the state of the art are compared in Table \ref{table:speed}. We achieve 75 fps for anomaly detection with a GeForce GTX TITAN X, faster than other state of the art methods with the same setting \cite{park2020learning}.

As can be seen in Fig. \ref{fig:recons-predict}, the future prediction of a biker becomes worse than the prediction of a pedestrian. The model is trained mostly on video sequences containing pedestrians, the prediction of the biker looks similar to the pedestrian. 
Here the prediction error is significantly larger than the reconstruction error. 

\vspace{-10pt}
\section{Conclusion}
We have adapted a state of the art predictive encoder-decoder deep network to detect abnormal events in video. We evaluate detection performance using both sequence prediction and reconstruction, and show that prediction gives superior performance on anomaly detection. For the prediction model, we obtain competitive performance to state of the art methods on five standard datasets. Finally, we evaluate performance across different prediction windows, encompassing varying levels of motion complexity. 
Our future work includes investigating the fusion of gray-scale images and optical flow on input. 

\ifCLASSOPTIONcaptionsoff \newpage\fi

\balance



\end{document}